\documentclass[final,3p,times]{elsarticle}

\usepackage{lineno,hyperref}
\usepackage{pifont}
\usepackage{amsmath}
\usepackage{amsfonts}
\usepackage{amssymb}
\usepackage{graphicx}
\usepackage{array}
\usepackage{booktabs}
\usepackage{float}
\usepackage{multirow}
\usepackage{color}
\usepackage[ruled]{algorithm2e}
\usepackage{subfigure}
\usepackage{bbding}
\usepackage{hyperref}
\usepackage{diagbox}
\usepackage{hyperref}
\usepackage{amsfonts,amssymb} 
\usepackage{soul, color, xcolor}
\usepackage{caption} 

\hypersetup{hidelinks,
	colorlinks=true,
	allcolors=black,
	pdfstartview=Fit,
	breaklinks=true}
\modulolinenumbers[5]
\biboptions{sort&compress}
\journal{Journal of \LaTeX\ Templates}

\begin{document}  

\begin{frontmatter}

\title{PAS-Mamba: Phase-Amplitude-Spatial State Space Model for MRI Reconstruction
}

\author[mymainaddress]{Xiaoyan Kui}
\author[mymainaddress]{Zijie Fan}
\author[mymainaddress]{Zexin Ji\corref{mycorrespondingauthor}}
\cortext[mycorrespondingauthor]{Corresponding Author}
\ead{zxjicsu@gmail.com}
\author[mysecondaddress]{Qinsong Li}
\author[mymainaddress]{Hao Xu}
\author[mythirdaddress]{Weixin Si}
\author[myforthaddress]{Haodong Xu}
\author[mymainaddress]{Beiji Zou}


\address[mymainaddress]{School of Computer Science and Engineering, Central South University, Changsha, 410083, China}
\address[mysecondaddress]{Big Data Institute, Central South University, Changsha, 410083, Hunan, China}
\address[mythirdaddress]{Shenzhen Institute of Advanced Technology, Chinese Academy of Science, Shenzheng, 518055, Guangdong, China}
\address[myforthaddress]{Department of Orthopaedics, The Second Xiangya Hospital of Central South University, Changsha, Hunan 410011, China}

\begin{abstract}

Joint feature modeling in both the spatial and frequency domains has become a mainstream approach in MRI reconstruction. However, existing methods generally treat the frequency domain as a whole, neglecting the differences in the information carried by its internal components. According to Fourier transform theory, phase and amplitude represent different types of information in the image. Our spectrum swapping experiments show that magnitude mainly reflects pixel-level intensity, while phase predominantly governs image structure. To prevent interference between phase and magnitude feature learning caused by unified frequency-domain modeling, we propose the Phase-Amplitude-Spatial State Space Model (PAS-Mamba) for MRI Reconstruction, a framework that decouples phase and magnitude modeling in the frequency domain and combines it with image-domain features for better reconstruction. In the image domain, LocalMamba preserves spatial locality to sharpen fine anatomical details. In frequency domain, we disentangle amplitude and phase into two specialized branches to avoid representational coupling. To respect the concentric geometry of frequency information, we propose Circular Frequency Domain Scanning (CFDS) to serialize features from low to high frequencies. Finally, a Dual-Domain Complementary Fusion Module (DDCFM) adaptively fuses amplitude–phase representations and enables bidirectional exchange between frequency and image domains, delivering superior reconstruction. Extensive experiments on the IXI and fastMRI knee datasets show that PAS-Mamba consistently outperforms state-of-the-art reconstruction methods.

\end{abstract}

\begin{keyword}
MRI reconstruction, State space models, Fourier domain, Cross-domain information fusion.
\end{keyword}

\end{frontmatter}

\section{Introduction}

Magnetic resonance imaging (MRI) has become one of the most important clinical diagnostic modalities because it offers non-invasive imaging, high spatial resolution, and excellent soft-tissue contrast. However, the long acquisition time often causes patient discomfort and may introduce motion-induced blurring, which limits its effectiveness in dynamic imaging and real-time applications. To shorten the scan duration, researchers usually acquire only a subset of k-space data through undersampling strategies, which can considerably accelerate the imaging process. Nevertheless, this approach inevitably introduces aliasing artifacts and degrades image quality~\cite{lustig2007sparse}. As a result, MRI reconstruction plays a vital role in recovering high-quality images from undersampled k-space data, ensuring a better trade-off between imaging speed and image fidelity~\cite{jaspan2015compressed,kim2024review,wang2025deep,qiu2025good}.
\begin{figure}[t]  
    \centering
    \includegraphics[width=0.7\textwidth]{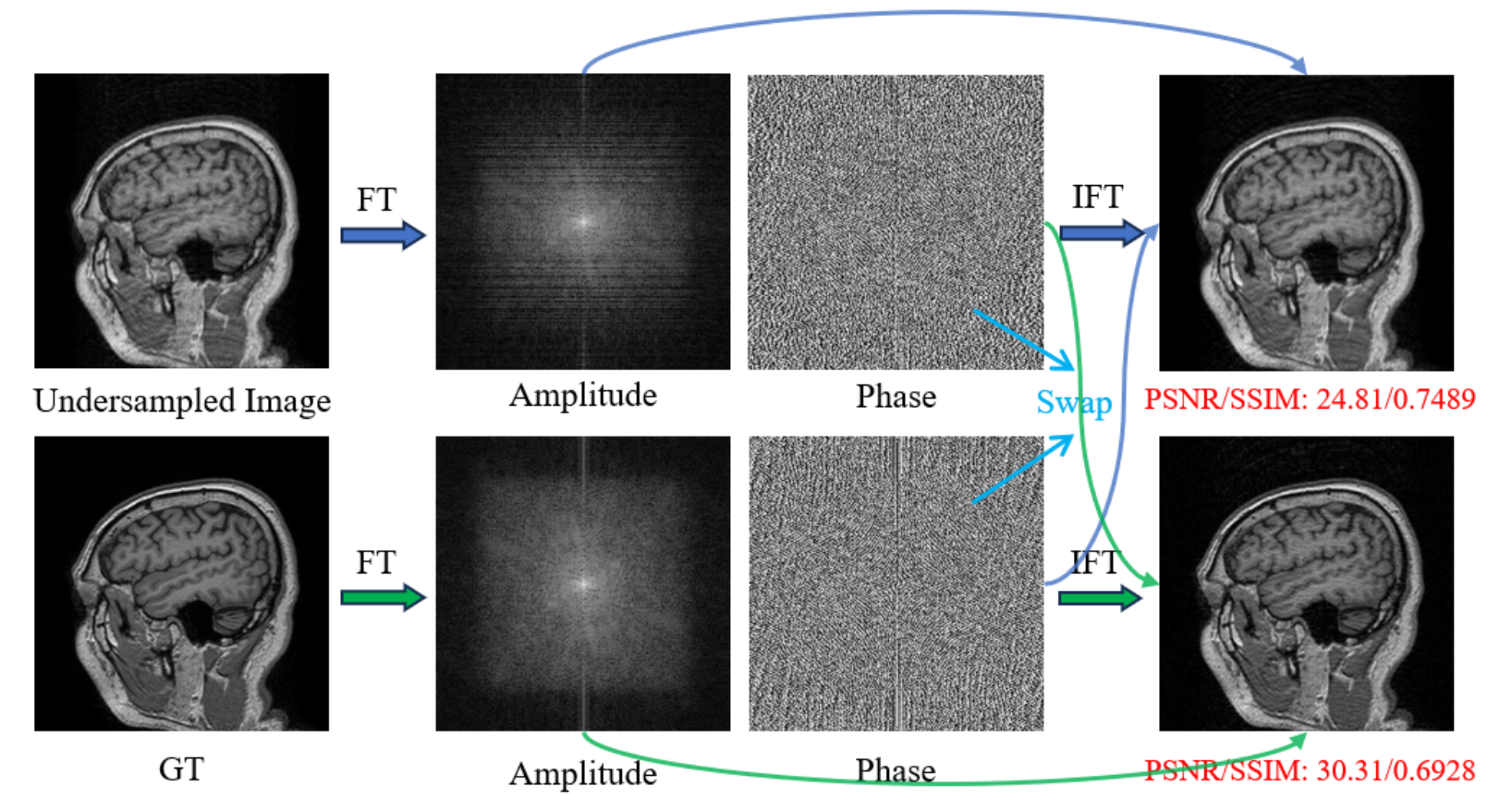}  
    \caption{Observation of the phase spectrum exchange.}  
    \label{fig:swap}  
\end{figure}
\begin{figure}[t]  
    \centering
    \includegraphics[width=0.88\textwidth]{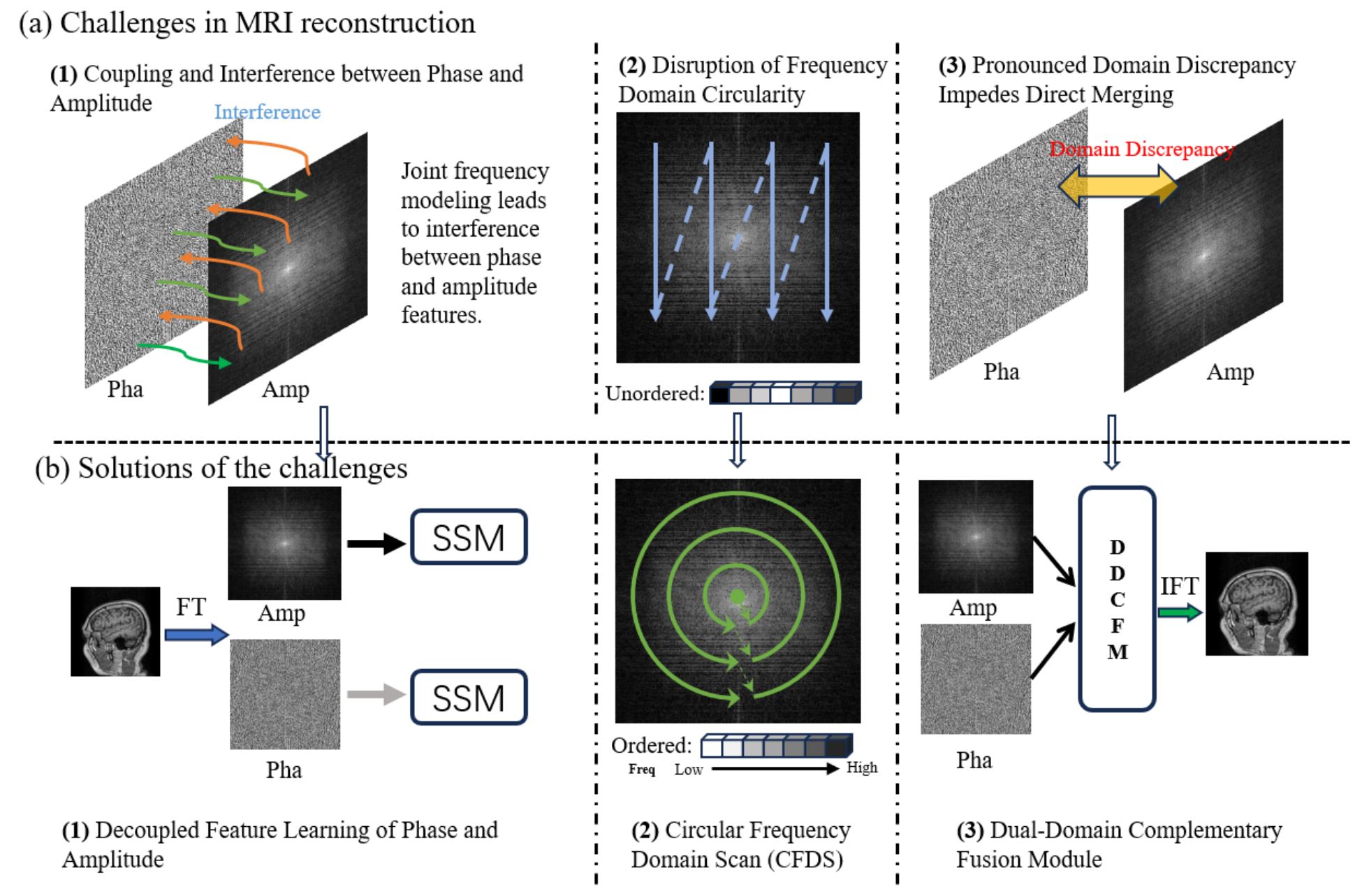}  
    \caption{Motivation: (a) Challenges in MRI reconstruction: the Coupling and Interference of Phase and Amplitude Information, along with the Disruption of the Circularity in the Frequency Domain and Significant Domain Discrepancies, Hinder Direct fusion. (b) To address these issues, we propose a decoupled modeling approach in the frequency domain, which separates the phase and amplitude components. Additionally, a circular scanning method is employed in the frequency domain to align with the concentric circular characteristics of the frequency domain. Furthermore, the Dual-Domain Complementary Fusion Module(DDCFM) is introduced to integrate the phase and amplitude information.  }  
    \label{fig:motivation}  
\end{figure}
The advent of deep learning techniques has led to the utilization of convolutional neural networks (CNNs) in many MRI reconstruction approaches for effective feature extraction~\cite{fang2025dugcn,qiao2025mcu,zou2022joint,wang2024crnn}. However, their limited local receptive fields hinder the modeling of long-range dependencies. This limitation is particularly pronounced in MRI reconstruction, since MRI data are originally acquired in the frequency domain~\cite{han2019k}, which exhibits inherent global characteristics and requires effective global modeling for accurate reconstruction~\cite{aggarwal2018modl}. To overcome this limitation, researchers have introduced Vision Transformers (ViTs) into image processing~\cite{huang2022swin,guo2023reconformer}. ViTs are capable of capturing long-range dependencies, and when integrated with convolutional blocks for local feature extraction, they have become a predominant approach in MRI reconstruction~\cite{hu2022trans,korkmaz2021deep}.
Despite their advantages, ViTs suffer from a notable drawback due to their quadratic computational complexity when capturing long-range dependencies.
To address this limitation, Mamba~\cite{gu2023mamba} emerges as a promising alternative by enabling global dependency modeling with linear computational complexity.
This design choice is particularly well aligned with MRI reconstruction, as MRI data are inherently acquired in k-space, where information is globally distributed and reconstruction quality is highly sensitive to effective global modeling.
Given Mamba’s strong capability for linear-complexity global representation learning~\cite{bansal2024comprehensive}, employing Mamba for feature extraction in MRI reconstruction is well motivated.

Grounded in the fundamental physics of MR image formation, we extend our investigation to the frequency domain. To further analyze the respective roles of amplitude and phase in MRI reconstruction, we conduct a phase spectrum exchange experiment on MRI images, as shown in Figure~\ref{fig:swap}. The results of the experiment indicated that preserving the fully-sampled amplitude led to a higher PSNR, while preserving the fully-sampled phase resulted in a higher SSIM value. It demonstrates that amplitude and phase encode distinct and complementary information: the phase spectrum primarily preserves structural details~\cite{van1999basic,cole2021analysis}, whereas the amplitude spectrum mainly reflects pixel-wise intensity variations. This observation motivates the separate modeling of amplitude and phase features in the frequency domain.

In addition to frequency-domain modeling, spatial-domain learning is also essential for preserving local structures in MRI reconstruction~\cite{liu2018applications}.
Frequency-domain features primarily encode global structural information, whereas local textures and fine anatomical details are more effectively captured in the image domain when spatial locality is preserved.
Therefore, incorporating an image-domain branch with locality-aware modeling provides complementary information to frequency-domain global representations.

 Nonetheless, applying Mamba to dual-domain MRI reconstruction faces three main challenges, as shown in Figure~\ref{fig:motivation}(a). (1) Modeling the entire frequency domain as a whole entangles the features of the phase and amplitude domains, causing mutual interference that impedes effective learning in either domain. (2) In the Fourier domain, components of the same frequency form concentric circles. Converting the two-dimensional feature map into a one-dimensional sequence using conventional scanning methods fails to preserve the inherent order of the features. (3) Significant discrepancies between the phase and amplitude domains limit the effectiveness of direct fusion in obtaining optimal feature representations.

To address the above challenges, we propose the Phase-Amplitude-Spatial
State Space Model (PAS-Mamba) for MRI Reconstruction. PAS-Mamba models the reconstruction process from complementary spatial- and frequency-domain perspectives and consists of a spatial Mamba branch for image-domain modeling and a frequency Mamba branch for frequency-domain modeling. In the image domain, we adopt a LocalMamba scanning strategy to preserve spatial locality, which enables more effective learning of local textures and fine-grained anatomical details. In the frequency domain shown in Figure~\ref{fig:motivation}(b), motivated by Fourier theory and phase spectrum exchange experiments, we decouple the frequency representation into amplitude and phase branches to avoid feature interference caused by unified modeling. Specifically, the amplitude branch mainly encodes intensity-related information, while the phase branch governs structural content. Moreover, to better align with the concentric characteristics of the frequency domain, we introduce a Circular Frequency Domain Scanning (CFDS) to unfold two-dimensional frequency features into ordered one-dimensional sequences from low to high frequencies. Finally, we design a Dual-Domain Complementary Fusion Module (DDCFM) to adaptively fuse phase and amplitude features and to facilitate effective information exchange between the frequency and spatial domains, leading to improved MRI reconstruction performance. 

Our contributions are summarized as follows:

\begin{itemize}[{\textbullet}]
\item We propose PAS-Mamba, a MRI reconstruction framework that decouples amplitude and phase in the frequency domain using Mamba, enabling domain-aware global dependency learning tailored to k-space characteristics, while also incorporating an image-domain branch to capture local spatial features.
\item We design a circular scanning strategy in the frequency domain to construct effective one-dimensional sequences, suited to the global properties of k-space. In the image domain, we adopt the LocalMamba scanning method to efficiently preserve local structures.
\item We introduce a Dual-Domain Complementary Fusion Module (DDCFM) to enable complementary fusion between phase and amplitude in the frequency domain and between the frequency and spatial domains for feature refinement.
\item Extensive experiments on IXI and fastMRI datasets demonstrate that PAS-Mamba consistently outperforms existing methods.
\end{itemize}

\section{Related Works}

\subsection{CNN-based Methods}
Traditional MRI reconstruction methods, including Compressed Sensing (CS) and Parallel Imaging(PI)~\cite{griswold2002generalized}, rely on iterative optimization and handcrafted priors. These approaches, however, tend to be computationally expensive, require careful parameter tuning, and are sensitive to noise. In contrast, convolutional neural networks (CNNs) have long been the dominant architecture for MRI reconstruction due to their strong ability to extract local features~\cite{chen2022pyramid,wang2024crnn,sun2025fourier,bongratz2024neural}. Early work, such as the U-Net~\cite{ronneberger2015u}, established a foundation for end-to-end mapping from undersampled to fully sampled MRI images. Its wide adoption in medical imaging has inspired numerous U-Net-based variants for MRI reconstruction.
For example, Zhou et al. proposed the Residual Non-Local Fourier Network (RNLFNet)~\cite{zhou2023rnlfnet}, which learns complementary information from both the image and k-space domains. Later, Yang et al. introduced the Artificial Fourier Transform Network (AFTNet)~\cite{yang2025deep}, which integrates complex-valued neural networks with the frequency domain to jointly process real and imaginary components of k-space. The Global Fourier Convolution Block (FCB)~\cite{sun2025fourier} was proposed to address the limited receptive field of CNN-based MRI reconstruction. By converting spatial convolutions into frequency-domain operations, FCB provides whole-image receptive fields with low computational cost and improves detail recovery.
Despite these advances, the convolutional operations have a restricted receptive field~\cite{liu2024image,sun2025fourier}, making it difficult to model long-range dependencies and capture global structures effectively.

\subsection{Transformer-based Methods}

To address the limitations of CNNs in capturing global dependencies, Vision Transformer (ViT) architectures have been introduced into MRI reconstruction. These models leverage self-attention mechanisms to effectively model long-range interactions across the image.
Guo et al. proposed a recurrent Transformer model called ReconFormer~\cite{guo2023reconformer}. It employs a recursive attention mechanism to utilize multi-scale information, improving reconstruction quality. Around the same time, Yi et al. introduced Frequency Learning via Multi-scale Fourier Transformer for MRI Reconstruction (FMTNet)~\cite{yi2023frequency}, which uses separate high-frequency and low-frequency branches to capture relevant frequency details. 
More recently, Chen et al. presented DDFCA-Net(a dual-path network with dual-domain fusion and cross-attention)~\cite{chen2025dual}, which enhances global feature learning by combining two parallel pathways: one based on CNNs for local features and another using a ViT for global features. By explicitly modeling both local and global information, many networks achieve impressive reconstruction performance~\cite{huang2022swin,shen2024magnetic,sheng2024cascade,al2024ocucformer}.
Despite these advances, Transformer-based methods typically involve high computational costs and substantial memory usage, which can hinder their practical adoption in clinical settings.

\subsection{State Space Models}
State space models(SSMs) have recently attracted considerable attention due to their ability to model global dependencies with linear computational complexity. Benefiting from this property, SSM-based Mamba has been widely adopted in medical imaging tasks such as medical image segmentation~\cite{ruan2024vm,DBLP:conf/icic/JiZKLVR25} and super-resolution~\cite{kui2025iterative,ji2025global,ji2025generation}, where it has demonstrated strong performance. MRI reconstruction is inherently sensitive to global modeling, as the original MRI data are acquired in k-space, which exhibits intrinsic global characteristics, and effective reconstruction often requires joint consideration of both frequency and image domains. Given its efficient global dependency modeling capability, Mamba is therefore particularly well suited for MRI reconstruction tasks~\cite{li2025lmo,zou2025mmr}.

In 2024, Guo et al. proposed MambaIR~\cite{guo2024mambair}, a simple baseline for image restoration based on the state-space model. It was the first work to apply Mamba in image restoration and fully exploits Mamba’s ability to capture global dependencies with linear complexity, providing an effective reference for subsequent MRI reconstruction research. In the same year, VM-UNet~\cite{ruan2024vm} was introduced, combining the strengths of Mamba and U-Net to leverage both U-Net’s multi-scale feature extraction and Mamba’s global dependency modeling.
Zou et al. proposed MMR-Mamba~\cite{zou2025mmr}, which performs reconstruction in both the image and frequency domains, effectively integrating multimodal features for MRI reconstruction. Also, Li et al. introduced the Linear Mamba Operator (LMO)~\cite{li2025lmo}. This method addresses the limited generalization of prior approaches under varying acceleration rates and maintains strong interpretability.

Nevertheless, existing Mamba-based MRI reconstruction methods still present several common limitations. Mamba modules are often adopted in a generic manner, while task-specific characteristics of MRI reconstruction, particularly frequency-domain priors, remain insufficiently exploited. Moreover, most approaches treat the frequency domain as a unified representation, without explicitly modeling the distinct structural roles of phase and amplitude, which may lead to suboptimal feature coupling. In addition, interactions between spatial and frequency domains are typically implemented through straightforward fusion strategies, which may introduce mutual interference. These observations suggest that, although Mamba provides an efficient mechanism for global dependency modeling, its potential in MRI reconstruction has not yet been fully realized.

\section{METHODOLOGY}
\subsection{Preliminarie}
The Mamba model~\cite{gu2023mamba} is based on the State Space Model (SSM)~\cite{gu2021combining}, which can capture state representations and forecast subsequent states. The conversion from a one-dimensional function or sequence $x(t) \in \mathbb{R}$ to an output $y(t) \in \mathbb{R}$ is facilitated by a hidden state $h(t) \in \mathbb{R}^{\mathbb{N}}$. This process is generally realized using linear ordinary differential equations (ODEs), as described below.

\begin{equation}\label{eq01}
    h^{\prime}(t)=\mathbf{A} h(t)+\mathbf{B} x(t), \quad y(t)=\mathbf{C} h(t).
\end{equation}
where $\mathbf{A} \in \mathbb{R}^{\mathrm{N} \times \mathrm{N}}$ is the state matrix, and $\mathbf{B} \in \mathbb{R}^{\mathrm{N} \times 1}$ and $\mathbf{C} \in \mathbb{R}^{1 \times \mathrm{N}}$ are the projection parameters. To adapt ODEs for deep learning, the zero-order hold (ZOH) method is employed for discretization. By introducing a timescale parameter $\Delta$, the continuous-time matrices $\mathbf{A}$ and $\mathbf{B}$ are converted into their discrete counterparts, $\overline{\mathbf{A}}$ and $\overline{\mathbf{B}}$, through the following steps:

\begin{equation}\label{eq2}
\overline{\mathbf{A}}=\exp (\Delta \mathbf{A}), \quad \overline{\mathbf{B}}=(\Delta \mathbf{A})^{-1}(\exp (\Delta \mathbf{A})-\mathbf{I}) \cdot \Delta \mathbf{B} .
\end{equation}

Following discretization, Eq.\ref{eq01} is reformulated for discrete-time processing as:

\begin{equation}\label{eq3}
h_t=\overline{\mathbf{A}} h_{t-1}+\overline{\mathbf{B}} x_t, \quad y_t=\mathbf{C} h_t .
\end{equation}

Mamba incorporates a selective mechanism that enables the model to dynamically adjust the state transition matrix in response to the input. Moreover, Mamba significantly enhances computational efficiency through optimized GPU memory access and computation processes. Compared to the quadratic complexity of Transformers, Mamba's linear complexity affords it a distinct advantage in handling long sequences.

\subsection{Overall architecture}
Let $I_{in}$ denote the undersampled image and $I_{gt}$ denote the fully-sampled image; their relationship is as follows:
\begin{equation}
I_{in}=\operatorname{F}^{-1}( M\odot \operatorname{F}(I_{gt}))+\varepsilon. 
\end{equation}
Here, $\operatorname{F}^{-1}$ and $\operatorname{F}$ denote the inverse and Fourier transforms, respectively, $\varepsilon$ signifies noise, and $M$ represents the undersampling mask. Our objective is to use $I_{in}$ as input to a neural network to achieve high-quality reconstruction result $I_{out}$. The MRI reconstruction problem can be expressed as an optimization issue as follows:
\begin{equation}
    I_{out} = argmin_{f(\theta )} \parallel I_{gt}-\operatorname{f}(I_{in});\theta \parallel_{1}.
\end{equation}

\begin{figure*}[t]  
    \centering
    \includegraphics[width=1.0\textwidth]{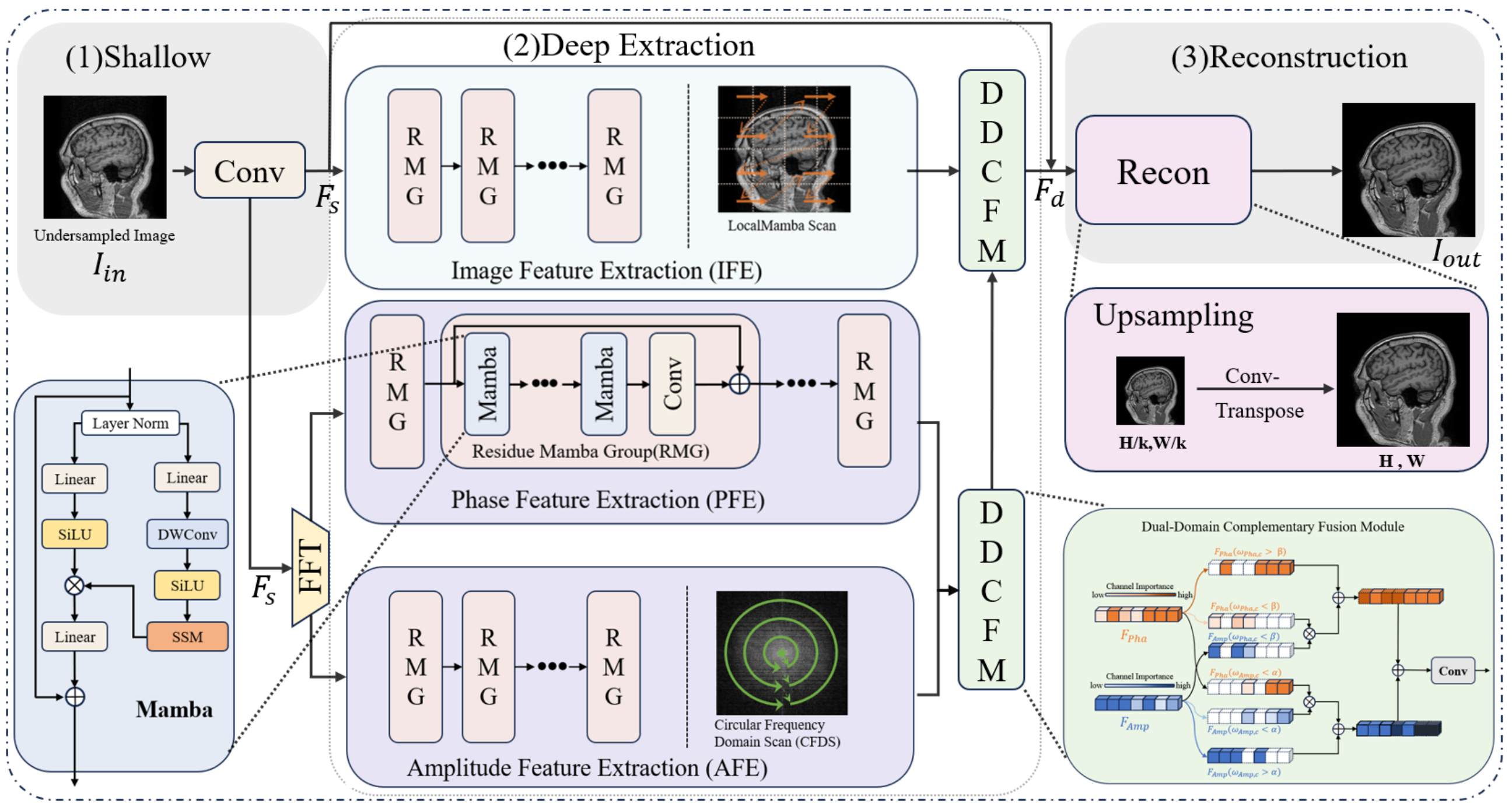}  
    \caption{The overall architecture of PAS-Mamba. Given an undersampled image($I_{in}$), the network splits the input into a spatial-domain branch and a frequency-domain branch. The frequency-domain branch is further divided into amplitude and phase sub-branches. The proposed DDCFM module first integrates the amplitude and phase features to form a complete frequency-domain representation, and then fuses this representation with the spatial-domain features. The fused features are finally fed into the reconstruction head to produce the final reconstructed image($I_{out}$).}  
    \label{fig:method}  
\end{figure*}

As illustrated in Figure~\ref{fig:method}, the main pipeline of our proposed PAS-Mamba can be divided into three stages.

(1)\textbf{Shallow extraction}: For the undersampled image input $I_{in}\in \mathbb{R}^{H\times W\times 1}  $, we apply a convolutional layer to obtain shallow features $F_{s}\in \mathbb{R}^{H\times W\times C}  $:
\begin{equation}
    F_{s}=\operatorname{Conv}(I_{in}).
\end{equation}

(2)\textbf{Deep extraction}: Subsequently, deep feature extraction is performed on $F_{s}$, which is split into two branches: the image domain and the frequency domain. In the image domain branch, features are extracted using the Image Feature Extraction (IFE) module, which consists of multiple Residue Mamba Group (RMG) modules. Each RMG module includs several Mamba blocks, and a convolutional block is appended at the end to enhance local feature learning. To better adapt to the characteristics of the image domain, we employ the localmamba scanning method to unfold the 2D data into 1D while preserving spatial locality.

In the frequency domain branch, $F_s$ is first transformed using the Fourier transform to extract phase and amplitude features. These features are then processed separately through the Phase Feature Extraction (PFE) and Amplitude Feature Extraction (AFE) modules. To preserve the concentric circular distribution characteristic of the frequency domain, we introduce the Circular Frequency Domain Scanning (CFDS), which arranges the frequencies from low to high. After obtaining deep phase and amplitude features, a Dual-Domain Complementary Fusion Module (DDCFM) is proposed to dynamically integrate the phase and amplitude features. Finally, the DDCFM was applied again to fuse the features from the image and frequency domain branches, resulting in the fused feature $F_d \in \mathbb{R}^{H \times W \times C}$.

(3)\textbf{Reconstruction}: After completing both shallow and deep feature extraction, the extracted features are fed into the reconstruction layer (Recon) to generate the reconstructed output $I_{\text{out}} \in \mathbb{R}^{H \times W \times 1}$:
\begin{equation}
    I_{out}=\operatorname{Recon}(F_{s},F_{d}).
\end{equation}
The reconstruction layer is composed of a transposed convolution module.

We employ a hybrid loss function to train our proposed PAS-Mamba:
\begin{equation}
L_{\rm total} = \|I_{\rm out}{-}I_{\rm gt}\|_{1} + \alpha\|P(I_{\rm out}){-}P(I_{\rm gt})\|_{1} + \beta\|A(I_{\rm out}){-}A(I_{\rm gt})\|_{1}.
\end{equation}
Here, $P$ represents the phase component, $A$ represents the amplitude component; the hybrid loss therefore supervises the image, magnitude, and phase domains, respectively. The $\alpha$ and $\beta$ are empirically set to 0.05. The use of this hybrid loss enables the network to jointly optimize reconstruction in both the spatial and frequency domains. The image-domain loss ensures pixel-level fidelity, while the phase and amplitude losses guide the model to better preserve the global structure and fine-grained details inherent in k-space. This combination helps achieve more accurate and visually faithful MRI reconstructions.

\subsection{LocalMamba Scanning}
\begin{figure}[t]  
    \centering
    \includegraphics[width=0.5\textwidth]{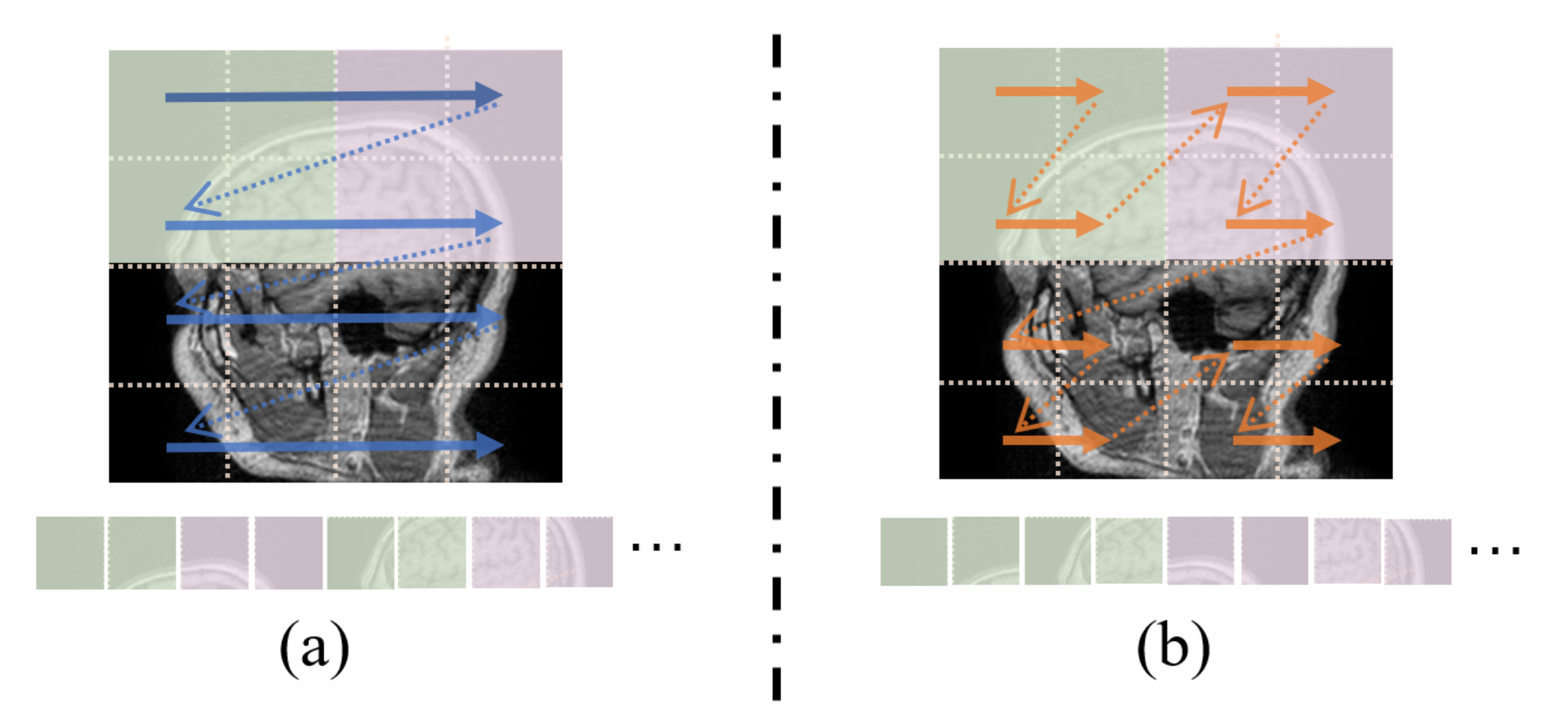}  
    \caption{(a) The standard Mamba scanning disrupts local spatial features in the image. (b) LocalMamba Scanning preserves local spatial features in the image. }  
    \label{fig:localmamba}  
\end{figure}

Local information is essential for preserving fine-grained structures in reconstructed images, which may not be fully captured by frequency-domain modeling alone. The image-domain branch in PAS-Mamba is designed to capture local spatial features, complementing the global dependency modeling in the frequency domain. Inspired by~\cite{huang2024localmamba}, we adopt the LocalMamba scanning strategy.  In Figure~\ref{fig:localmamba}(a), the conventional row-wise scanning method is shown, which disrupts the local neighborhood relationships of image patches. In contrast, Figure~\ref{fig:localmamba}(b) presents the LocalMamba scanning pattern, where each image patch is traversed along a local pattern that preserves neighborhood relationships. These sequences are then fed into the image-domain Mamba branch. By using LocalMamba scanning, the image-domain branch complements the frequency-domain modeling, allowing PAS-Mamba to leverage both local and global information effectively.

\subsection{Circular Frequency Domain Scanning}

\begin{figure}[t]  
    \centering
    \includegraphics[width=0.7\textwidth]{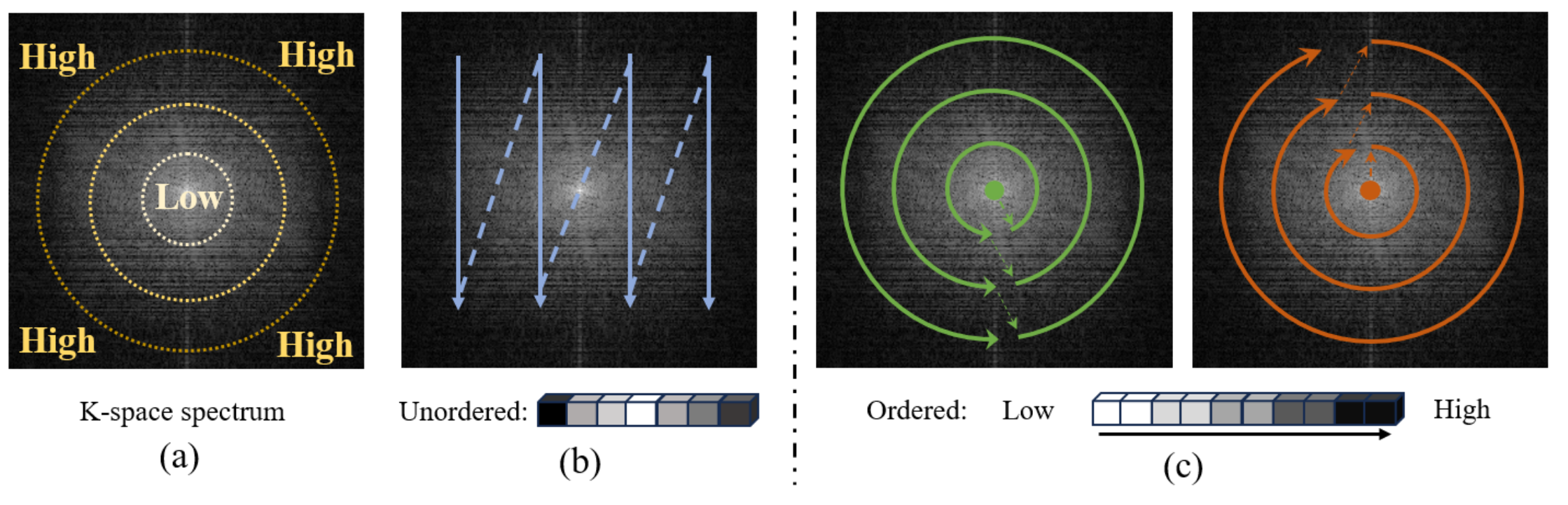}  
    \caption{(a) In the K-space spectrum, the frequency distribution exhibits a concentric circular structure. Low-frequency components are located at the center, while high-frequency components lie in the peripheral regions. (b) The standard Mamba scanning disrupts the concentric circular structure of the frequency domain. (c) Circular Frequency Domain Scanning (CFDS). This scanning method can generate ordered sequences from low to high frequencies. }  
    \label{fig:yuan_scan}  
\end{figure}
In Vanilla Mamba~\cite{liu2024vmamba}, a 2D feature map is flattened into a 1D vector along four directions (top-left, bottom-right, bottom-left, and top-right) using row-wise or column-wise operations, as depicted in Figure~\ref{fig:yuan_scan}(b). However, this scanning strategy does not consider the frequency distribution of k-space. Typically, k-space data are centered so that low-frequency components occupy the image center, while high-frequency components appear near the four corners, as depicted in Figure~\ref{fig:yuan_scan}(a). Components with the same frequency lie along concentric circles. As a result, standard Mamba scanning cannot produce a sequence ordered from low to high frequencies.
To address this limitation, we propose a novel scanning method, the Circular Frequency Domain Scanning (CFDS), as illustrated in Figure~\ref{fig:yuan_scan}(c). This method acquires components sequentially from low to high frequencies along circular paths. According to our strategy, there are eight possible scanning paths, determined by four starting points (top-left, bottom-left, top-right, and bottom-right) and two directions (clockwise and counterclockwise). To reduce redundancy, we select only four of these paths: top-left clockwise, top-right counterclockwise, bottom-left counterclockwise, and bottom-right clockwise. This approach preserves the concentric circular organization inherent to k-space, where coefficients at the same radial distance correspond to similar spatial frequency components. By unfolding the frequency domain along circular trajectories, the proposed scanning strategy maintains a physically meaningful ordering from low to high frequencies and avoids the disruption of frequency continuity introduced by conventional row-wise or column-wise scanning. This ordered representation enables coherent information propagation across different frequency bands, allowing the model to capture both low-frequency global structures and high-frequency fine details more effectively. Consequently, the learned representations align more closely with the intrinsic frequency characteristics of MRI data.

\subsection{Dual-Domain Complementary Fusion Module}
\begin{figure}[t]  
    \centering
    \includegraphics[width=0.6\textwidth]{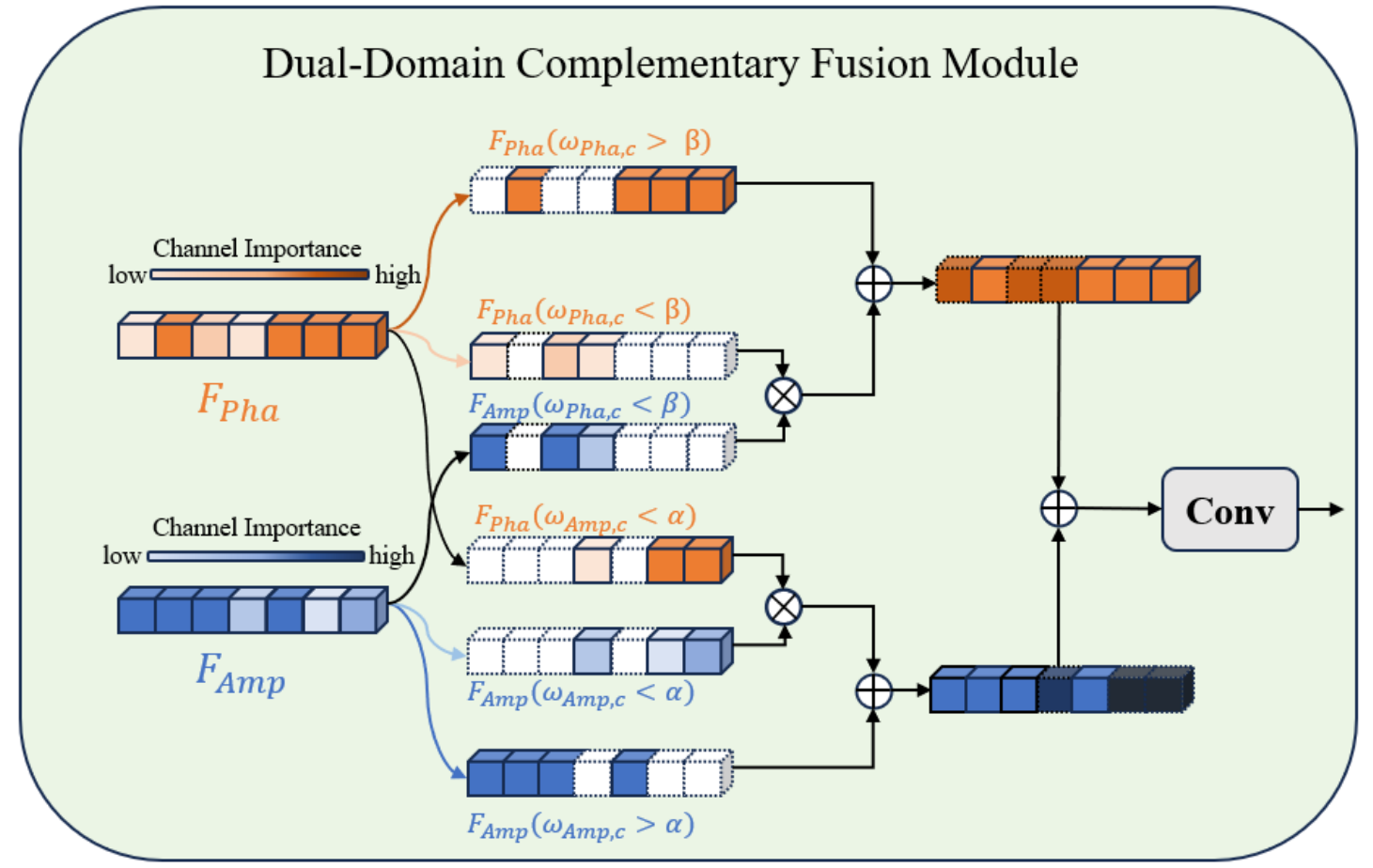}  
    \caption{Dual-Domain Complementary Fusion Module (DDCFM).}  
    \label{fig:DDCFM}  
\end{figure}

After extracting features from the amplitude and phase branches using our designed AFE and PFE modules, resulting in $F_{\text{Amp}}$ and $F_{\text{Pha}}$, traditional methods often apply a Fourier transform directly on these features to obtain image-domain data~\cite{li2024fouriermamba,zou2024freqmamba}. However, this direct fusion approach has clear limitations, as it overlooks the significant domain discrepancy between the amplitude and phase branches. Since the two domains capture different types of information, directly merging their features can mask useful information and fail to exploit their complementary properties effectively.
It is important to note that the corresponding positions in both spectra represent the same frequency components. This property allows the two domains to be linked through the same frequency within the same channel. Based on this observation, we introduce the Dual-Domain Complementary Fusion Module (DDCFM) to overcome the limitations of direct fusion and fully leverage the complementarity between amplitude and phase. The specific design of the DDCFM is illustrated in Figure~\ref{fig:DDCFM}. Using this module, channels with insufficient information in one domain can utilize corresponding channel data from the other domain to enhance themselves. This process maximizes the utilization of complementary information between the two domains.

Specifically, for a multi-channel feature, we assign an importance weight to each channel to quantify the amount of information it contains. Channels with less information are scaled down, while those with more information are preserved as much as possible. This idea can be effectively implemented using Instance Normalization (IN). The formula for Instance Normalization is as follows:
\begin{equation}
\operatorname{IN}(F_{Pha,c} )=\omega _{Pha,c} \frac{F_{Pha,c}-\mu _{Pha,c} }{\sqrt{\sigma_{Pha,c} ^{2}+\varepsilon  } }+\gamma _{Pha,c} \label{eq:1} 
\end{equation}
\begin{equation}
\operatorname{IN}(F_{Amp,c} )=\omega _{Amp,c} \frac{F_{Amp,c}-\mu _{Amp,c} }{\sqrt{\sigma_{Amp,c} ^{2}+\varepsilon  } }+\gamma _{Amp,c} \label{eq:2} 
\end{equation}
where $\omega$ denotes the channel importance weight. $\gamma _{Pha,c}$ and $\gamma _{Amp,c}$ are offsets, and $\varepsilon$ is a small constant to avoid division by zero.

After the channel importance weight $\omega$ is trained, we select the data from the corresponding channel in the other domain to enhance it. The specific implementation is as follows:

\begin{equation}
Enhanced_{Pha,c} = 
\begin{cases}
F_{Pha,c} \otimes F_{Amp,c}, & \text{if } \omega_{Pha,c} < \beta; \\
F_{Pha,c}, & \text{if } \omega_{Pha,c} \geq \beta.
\end{cases}\label{eq:3}
\end{equation}
\begin{equation}
Enhanced_{Amp,c} = 
\begin{cases}
F_{Amp,c} \otimes F_{Pha,c}, & \text{if } \omega_{Amp,c} < \alpha ; \\
F_{Amp,c}, & \text{if } \omega_{Amp,c} \geq \alpha.
\end{cases}\label{eq:4}
\end{equation}
Herein, $\otimes$ denotes element-wise multiplication, while $\alpha$  and $\beta$ represent the channel importance thresholds for the amplitude and phase domains, respectively. Channels with importance values below these thresholds are deemed to have insufficient information and thus require enhancement using data from the counterpart domain. We specifically select the bottom 10\% of channels with the least information as those to be enhanced. The formulas for calculating $\alpha$ and $\beta$ are as follows:
\begin{equation}
\alpha = \operatorname{TopK}(\omega_{Amp}, 0.1),
\label{eq:5}
\end{equation}
\begin{equation}
\beta = \operatorname{TopK}(\omega_{Pha}, 0.1),
\label{eq:6}
\end{equation}

Herein, $\omega _{Amp}$ and $\omega _{Pha}$ denote the channel importance weights for the amplitude and phase domains, respectively. TopK signifies the operation of selecting the bottom 10\% of channels with the smallest weights.

In the DDCFM module, phase and amplitude features fully leverage complementary information from the counterpart domain to enhance their respective feature representations, followed by convolutional fusion. Moreover, the same fusion mechanism is extended to integrate frequency-domain and spatial-domain features, enabling complementary interaction between global frequency information and local spatial structures.

\section{Experiments and results}
\subsection{Dataset description}
In this study, we employed two publicly available datasets to experimentally evaluate our methods: the IXI dataset and the fastMRI dataset. 

\textbf{The IXI Dataset}\footnote{https://brain-development.org/ixi-dataset/} encompasses a variety of MRI contrast types, including T1-weighted, T2-weighted, Proton Density (PD)-weighted, and Magnetic Resonance Angiography (MRA) images. In this work, we exclusively use T1-weighted images for all experiments. From the IXI dataset, we randomly selected 465 volumes, which were then partitioned into training and testing sets at a ratio of 3:1. Among these samples, we obtained a total of 1,396 slices for the training set and 464 slices for the testing set. The size of each 2D image was 256 × 256 pixels. Notably, to simulate motion artifacts that occur due to patient movement during actual MRI acquisition, we applied random noise to the slices after applying an undersampling mask.

\textbf{The fastMRI Dataset}\footnote{https://fastmri.med.nyu.edu/} comprises MRI scans of the knee and brain, offering raw k-space data, multi-coil signals, and magnitude images. We adopt the T1-weighted knee subset in our experiments. We randomly selected 200 volumes from the fastMRI dataset, which were then divided into training and testing sets at a ratio of 4:1. In these samples, we acquired a total of 1,600 slices for the training set and 400 slices for the testing set. The size of each 2D image was 320 × 320 pixels.

\begin{table*}[t]
    \centering
    \caption{The numerical results on IXI under two masks with $\times2$ and $\times4$ acceleration ratios.}
    \label{tab:ixi_methods}
    \setlength{\tabcolsep}{12pt}      
\renewcommand{\arraystretch}{1.05} 
    \begin{tabular}{llllll}
    \hline
    & & \multicolumn{4}{c}{IXI} \\ \cline{3-6}
    AR & Methods & \multicolumn{2}{c}{Radial} & \multicolumn{2}{c}{Cartesian} \\ \cline{3-6}
    & & PSNR($\uparrow$) & SSIM($\uparrow$) & PSNR($\uparrow$) & SSIM($\uparrow$) \\ \hline
    \multirow{7}{*}{$\times2$}
    & ZF            & 29.67 & 0.6583 & 25.16 & 0.7478 \\
    & UNet          & 38.85 & 0.9713 & 34.52 & 0.9537 \\
    & SwinIR        & 40.01 & 0.9801 & 35.35 & 0.9595 \\
    & VM-UNet       & 38.91 & 0.9797 & 33.83 & 0.9541 \\
    & MambaIR       & 40.03 & 0.9769 & 36.11 & 0.9641 \\
    & FCB-UNet      & 40.03 & 0.9543 & 35.53 & 0.9461 \\
    & \textbf{Ours} & \textbf{40.43} & \textbf{0.9805} & \textbf{36.48} & \textbf{0.9694} \\ \hline
    \multirow{7}{*}{$\times4$}
    & ZF            & 21.72 & 0.4678 & 19.80 & 0.6001 \\
    & UNet          & 33.66 & 0.9304 & 29.15 & 0.8877 \\
    & SwinIR        & 34.48 & 0.9515 & 29.91 & 0.8978 \\
    & VM-UNet       & 32.99 & 0.9451 & 26.73 & 0.8423 \\
    & MambaIR       & 34.57 & 0.9461 & 29.70 & 0.8869 \\
    & FCB-UNet      & 34.18 & 0.9030 & 29.03 & 0.7828 \\
    & \textbf{Ours} & \textbf{35.11} & \textbf{0.9572} & \textbf{31.01} & \textbf{0.9197} \\ \hline
    \end{tabular}
\end{table*}

\begin{table*}[t]
    \centering
    \caption{The numerical results on fastMRI under two masks with $\times2$ and $\times4$ acceleration ratios.}
    \label{tab:fastmri_methods}
    \setlength{\tabcolsep}{10pt}      
\renewcommand{\arraystretch}{1.05} 
    \begin{tabular}{llllll}
    \hline
    & & \multicolumn{4}{c}{fastMRI} \\ \cline{3-6}
    AR & Methods & \multicolumn{2}{c}{Radial} & \multicolumn{2}{c}{Cartesian} \\ \cline{3-6}
    & & PSNR($\uparrow$) & SSIM($\uparrow$) & PSNR($\uparrow$) & SSIM($\uparrow$) \\ \hline
    \multirow{7}{*}{$\times2$}
    & ZF            & 29.71 & 0.8041 & 23.08 & 0.6762 \\
    & UNet          & 32.23 & 0.8796 & 28.82 & 0.8216 \\
    & SwinIR        & 32.12 & 0.8817 & 28.27 & 0.8109 \\
    & VM-UNet       & 32.11 & 0.8803 & 28.06 & 0.7990 \\
    & MambaIR       & 32.30 & 0.8807 & 28.81 & 0.8185 \\
    & FCB-UNet      & 31.98 & 0.8813 & 28.27 & 0.8264 \\
    & \textbf{Ours} & \textbf{32.46} & \textbf{0.8838} & \textbf{28.84} & \textbf{0.8301} \\ \hline
    \multirow{7}{*}{$\times4$}
    & ZF            & 25.43 & 0.6372 & 21.789 & 0.5611 \\
    & UNet          & 28.12 & 0.7032 & 25.95  & 0.6916 \\
    & SwinIR        & 28.13 & 0.7031 & 26.07  & 0.6808 \\
    & VM-UNet       & 27.27 & 0.6697 & 25.83  & 0.6629 \\
    & MambaIR       & 28.29 & 0.7001 & 26.28  & 0.7062 \\
    & FCB-UNet      & 27.72 & 0.6982 & 26.16  & 0.6801 \\
    & \textbf{Ours} & \textbf{28.35} & \textbf{0.7058} & \textbf{26.47} & \textbf{0.7081} \\ \hline
    \end{tabular}
\end{table*}

\subsection{Experimental setup}
\textbf{Comparison Methods.} To thoroughly evaluate the performance of our proposed model, we compared it with representative MRI reconstruction methods: ZF~\cite{bernstein2001effect}, UNet~\cite{ronneberger2015u}, SwinIR~\cite{swinir}, VM-UNet~\cite{ruan2024vm}, MambaIR~\cite{guo2024mambair} , and FCB-UNet~\cite{sun2025fourier}. For fairness, we adopted the training configurations reported in their original papers.

\textbf{Performance Metrics.} For quantitative evaluation, we report peak signal-to-noise ratio (PSNR) and structural similarity index (SSIM)~\cite{wang2003multiscale}. For qualitative analysis, we visualize the reconstruction results using error maps.

\textbf{Implementation Details.} The proposed model was implemented in PyTorch and trained on an RTX A6000 GPU. We employed the Adam optimizer with an initial learning rate of 0.0001. For each dataset, both radial and Cartesian undersampling masks were used during training.

\subsection{Experimental results}
\textbf{Quantitative Results.} In Table~\ref{tab:ixi_methods} and Table~\ref{tab:fastmri_methods}, we report the PSNR and SSIM results for all models across both datasets under ×2 and ×4 acceleration. 
Across different undersampling masks, radial sampling yields better reconstruction performance than Cartesian sampling. This is mainly because radial acquisition densely samples the central k-space region, which preserves low-frequency components encoding global structures and intensity distributions. As a result, the model can focus more on recovering high-frequency details. In contrast, Cartesian sampling causes substantial loss of both low- and high-frequency information, making it more challenging to jointly reconstruct global structures and local details.

For different datasets, we observed that reconstruction metrics on the IXI dataset are consistently higher than those on the fastMRI dataset. This difference can be attributed to several factors. First, the image sizes differ: IXI images are 256×256 brain MRIs, whereas fastMRI images are 320×320 knee MRIs. With the same acceleration factor, more data points need to be reconstructed in fastMRI, making the task more challenging. Second, the anatomical structures differ: brain MRIs are relatively regular and homogeneous, while knee MRIs are more complex with abundant details. These factors explain why the same model achieves higher PSNR and SSIM on IXI than on fastMRI. With respect to acceleration factors, reconstruction performance under ×4 is significantly lower than under ×2, as higher acceleration inevitably results in more missing k-space samples and increased reconstruction difficulty.

In our experiments, PAS-Mamba consistently achieved the best performance across different datasets and undersampling masks. Notably, under the IXI radial ×2 setting, it achieves a PSNR of 40.43 dB and an SSIM of 0.9805. Unlike most existing methods that treat the frequency domain as a unified representation, PAS-Mamba explicitly decouples phase and amplitude modeling, which better aligns with the intrinsic characteristics of k-space. Moreover, by combining Mamba-based global dependency modeling in the frequency domain with locality-preserving LocalMamba in the image domain, PAS-Mamba effectively balances global structure recovery and local detail preservation. In addition, the proposed Dual-Domain Complementary Fusion Module (DDCFM) enables more effective integration of spatial- and frequency-domain features, reducing feature interference and further improving reconstruction quality. Beyond these general trends, the superior performance of PAS-Mamba can be attributed to its key design choices. 
\begin{figure*}[t]  
    \centering
    \includegraphics[width=1.0\textwidth]{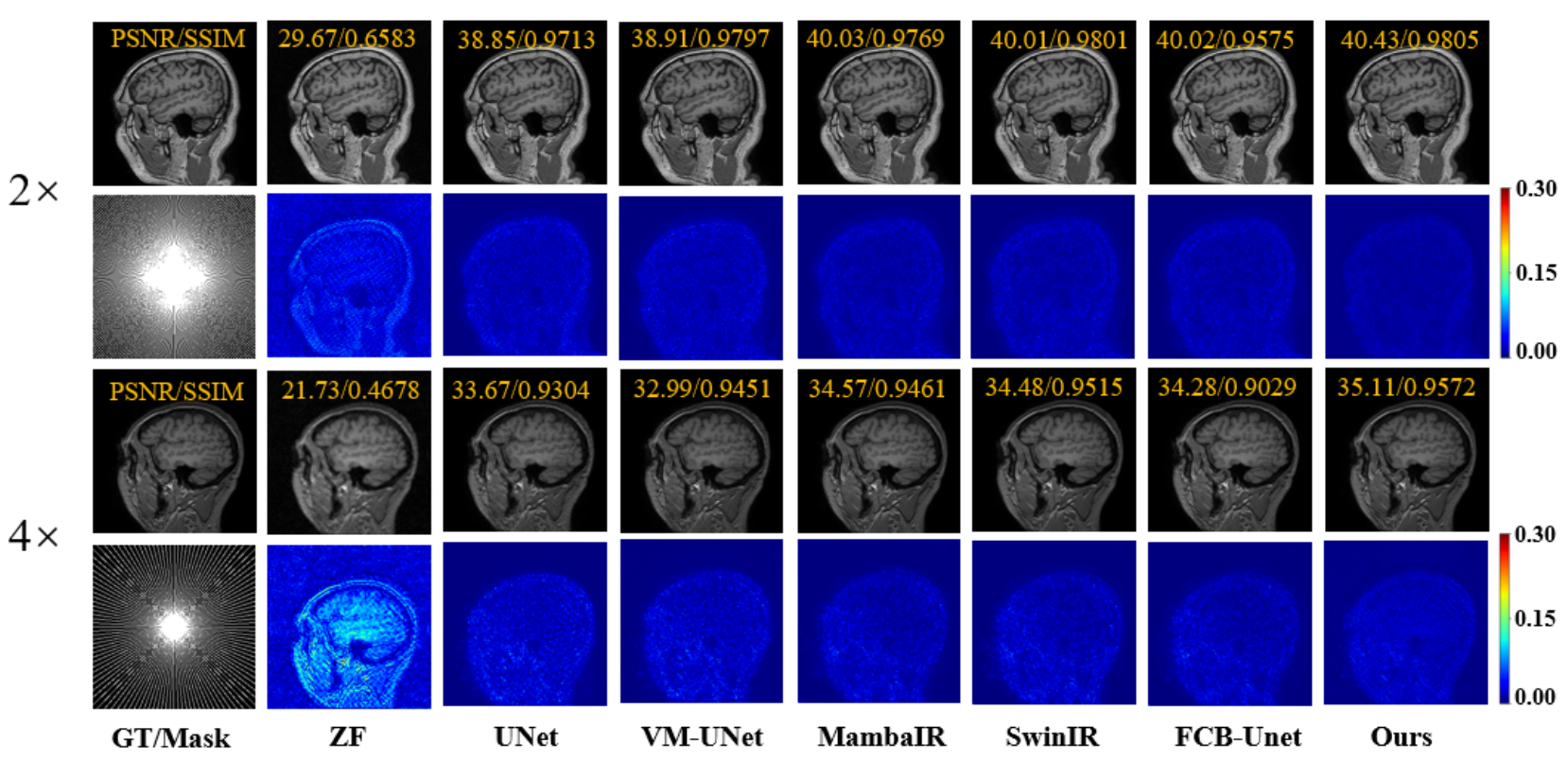}  
    \caption{Visual results and error maps on IXI dataset with radial mask ×2 and radial mask ×4 ARs.}  
    \label{fig:result_IXI}  
\end{figure*}

\begin{figure*}[t]  
    \centering
    \includegraphics[width=1.0\textwidth]{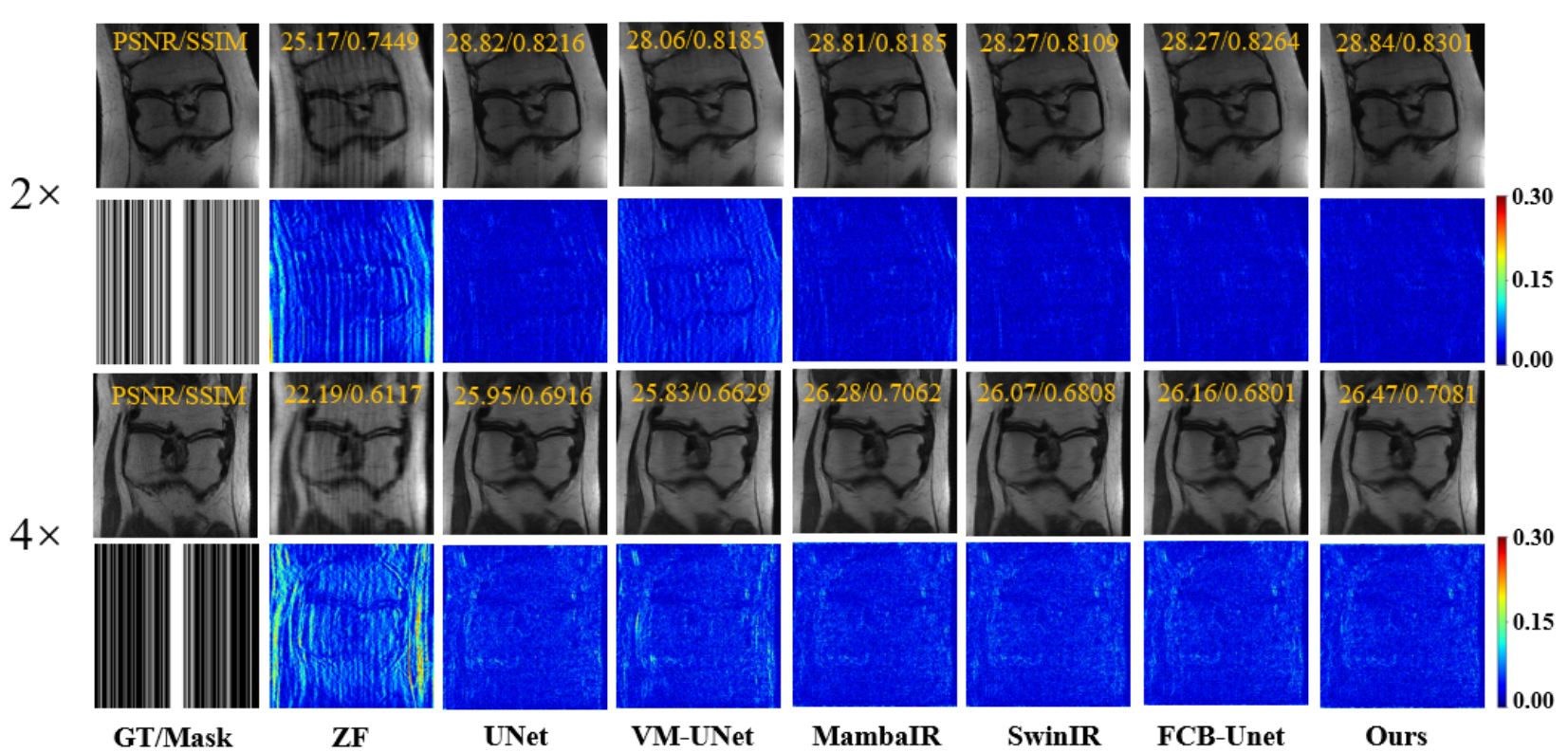}  
    \caption{Visual results and error maps on fastMRI dataset with radial mask ×2 and radial mask ×4 ARs.}  
    \label{fig:result_fastMRI}  
\end{figure*}

\textbf{Qualitative Results.} To further complement the quantitative results, Figure~\ref{fig:result_IXI} and Figure~\ref{fig:result_fastMRI} present visual comparisons of different methods under 2× and 4× acceleration on the IXI and fastMRI datasets, respectively. In the error maps, blue regions indicate small errors, while red regions highlight areas with larger reconstruction differences. The error maps show that zero-filled (ZF) reconstructions suffer from severe artifacts, which become more pronounced as the acceleration factor increases. 
Compared with ZF, other learning-based methods significantly reduce reconstruction errors, indicating their ability to suppress undersampling artifacts. 
Nevertheless, these methods still exhibit more residual errors than our approach. 
It is important to emphasize that this improvement does not stem from ignoring the frequency domain in prior methods, but rather from placing greater emphasis on its internal structure. 
Given that MRI data are intrinsically acquired in k-space, reconstruction performance is highly sensitive to how frequency-domain information is modeled. 
Instead of treating the frequency domain as a unified representation, our method explicitly distinguishes and separately models amplitude and phase components, which encode complementary information.
This design better aligns with the physical characteristics of MRI acquisition and enables more effective suppression of structured undersampling artifacts, leading to more accurate and stable reconstructions.

\subsection{Ablation studies}

To validate the contribution of each proposed component, we conduct ablation studies on the IXI dataset under a ×2 acceleration setting. The evaluated configurations are as follows:
(a) Baseline: a baseline model without any of the proposed components;
(b) w/o LocalMamba: our PAS-Mamba model without the LocalMamba scanning strategy;
(c) w/o CFDS: PAS-Mamba without the Circular Frequency Domain Scan(CFDS), where the Vanilla Mamba scanning strategy is used instead;
(d) w/o DDCFM: PAS-Mamba without the Dual-Domain Complementary Fusion Module(DDCFM) module;
(e) Phase-Amplitude-Spatial State Space Model for MRI Reconstruction(PAS-Mamba): the full PAS-Mamba model with all proposed components included.
The results are presented in Table~\ref{tab:ablation1}. The results indicate that the removal of any single component causes a clear drop in performance, which shows that each improvement plays an essential role. When all three components are incorporated into the baseline, the performance increases significantly. 

\begin{table*}[t]
    \centering
    \caption{Ablation study of the proposed modules on IXI dataset.}
    \setlength{\tabcolsep}{10pt}      
\renewcommand{\arraystretch}{1.05} 
\begin{tabular}{lcccllll}
\hline
Model & LocalMamba & CFDS & DDCFM & \multicolumn{2}{l}{Radial(×2)}       & \multicolumn{2}{l}{Cartesian(×2)}      \\ \cline{5-8} 
      &            &      &       & PSNR($\uparrow$)           & SSIM($\uparrow$)            & PSNR($\uparrow$)             & SSIM($\uparrow$)            \\ \hline
Baseline     & \ding{55}           & \ding{55}     & \ding{55}      & 39.71          & 0.9776          & 35.37            & 0.9621          \\
w/o LocalMamba     & \ding{51}          & \ding{51}    & \ding{55}      & 40.03          & 0.9791          & 36.41          & 0.9661           \\
w/o CFDS     & \ding{51}          & \ding{55}     & \ding{51}    & 40.14          & 0.9776          & 36.42          & 0.9644          \\
w/o DDCFM     & \ding{55}          & \ding{51}    & \ding{51}     & 40.06          & 0.9801          & 36.06          & 0.9641          \\
PAS-Mamba     & \ding{51}         & \ding{51}   & \ding{51}     & \textbf{40.43} & \textbf{0.9805} & \textbf{36.48} & \textbf{0.9694} \\ \hline
\end{tabular}
    \label{tab:ablation1}
\end{table*}

\subsubsection{Analysis of the LocalMamba Scanning}

We conducted ablation experiments for the LocalMamba scanning strategy in the image-domain branch, with results shown in Table~\ref{tab:ablation1}. The model using LocalMamba achieves an average improvement of 0.4 dB in PSNR and 0.0014 in SSIM compared with the w/o LocalMamba variant, indicating that LocalMamba better preserves the spatial locality of image patches and enables more effective extraction of local features. When combined with frequency-domain features capturing global dependencies, this complementary information contributes to improved overall reconstruction performance.
These findings confirm that LocalMamba is particularly suitable for the image-domain branch, enhancing local feature extraction while synergizing with frequency-domain global modeling.

\subsubsection{Analysis on the Circular Frequency Domain Scanning(CFDS)}

The CFDS is designed to adapt to the characteristics of the frequency domain. By unfolding 2D frequency-domain features into ordered 1D sequences from low to high frequencies, CFDS preserves frequency consistency and enables more effective modeling of global k-space dependencies. The w/o CFDS model result in Table~\ref{tab:ablation1} demonstrates the effectiveness of the proposed CFDS. In PAS-Mamba, the 2D frequency-domain features are unfolded along four directional scanning paths to capture directional dependencies in k-space(top-left clockwise, top-right counterclockwise, bottom-left counterclockwise, and bottom-right clockwise).

To further investigate the impact of different scanning strategies, we conduct ablation experiments with varying numbers of scanning directions, as summarized in Table~\ref{tab:ablation_cdfs}. Specifically, the 1-direction setting corresponds to a top-left clockwise scan, starting from the top-left corner and proceeding in a clockwise manner. The results show that the four-direction scheme consistently outperforms the single-direction setting, indicating that multi-directional unfolding enables more comprehensive modeling of frequency-domain dependencies.
When increasing the number of scanning directions from four to eight, only marginal performance gains are observed. Although additional directions allow the model to capture feature variations from more orientations, they also introduce redundant information and increase computational cost, leading to diminishing returns. Considering both reconstruction performance and computational efficiency, we adopt the four-direction CFDS strategy in the final model, which provides a well-balanced and practical design choice.
\begin{table}[t]
    \centering
    \caption{Ablation experiments for different design choices of CFDS.(IXI-Radial×2)}
    \setlength{\tabcolsep}{20pt}      
\renewcommand{\arraystretch}{1.05}
\begin{tabular}{lll}
\hline
settings                   & PSNR($\uparrow$)  & SSIM($\uparrow$)    \\ \hline
1-direction & 40.12 & 0.9759  \\ 
4-direction              & 40.43 & 0.9805\\ 
8-direction                 & 40.45 & 0.9811 \\ \hline
\end{tabular}
\label{tab:ablation_cdfs}
\end{table}
\begin{table}[t]
    \centering
    \caption{Ablation experiments for different design choices of DDCFM.(IXI-Radial×2)}
    \setlength{\tabcolsep}{20pt}      
\renewcommand{\arraystretch}{1.05}
\begin{tabular}{lll}
\hline
settings                   & PSNR($\uparrow$)  & SSIM($\uparrow$)   \\ \hline
w/o DDCFM-AP and DDCFM-IF & 40.03 & 0.9791 \\ 
w/o DDCFM-IF              & 40.08 & 0.9795 \\ 
PAS-Mamba                    & \textbf{40.43} &\textbf{0.9805} \\ \hline
\end{tabular}
\label{tab:ablation_ddcfm}
\end{table}
\subsubsection{Analysis on the Dual-Domain Complementary Fusion Module(DDCFM)}
The DDCFM is designed to exploit complementary information between domains, aiming to enhance feature integration and improve reconstruction performance. Specifically, DDCFM-AP represents the fusion of the amplitude and phase domains, whereas DDCFM-IF represents the fusion of the image and frequency domains. Ablation studies were conducted on the IXI dataset to validate the efficacy of DDCFM in fusing two domains. The experimental outcomes are depicted in Table~\ref{tab:ablation_ddcfm}. The experimental results demonstrate that the baseline model incorporating DDCFM-AP and DDCFM-IF attains the optimal reconstruction performance. This illustrates that the DDCFM module is effective not only for the fusion of the amplitude and phase domains but also for other types of bimodal domain fusions. The results indicate that DDCFM effectively integrates different types of dual-domain features, suggesting its potential applicability to other complementary domain fusion tasks.

\section{Conclusion and future work}
In this work, we propose PAS-Mamba, a MRI reconstruction framework that jointly considers spatial- and frequency-domain representations through a decoupled phase–amplitude modeling strategy. By explicitly recognizing the complementary roles of phase and amplitude information in k-space, PAS-Mamba constructs independent yet cooperative feature representations and integrates them with spatial-domain features to achieve high-quality reconstruction. To this end, PAS-Mamba leverages Mamba-based state space modeling to enable efficient global representation learning with linear computational complexity, employs a locality-aware LocalMamba strategy in the image domain to enhance local structural modeling, and introduces a circular frequency-domain scanning scheme that converts 2D k-space features into an ordered 1D sequence aligned with the intrinsic concentric structure of k-space. Furthermore, a Dual-Domain Complementary Fusion Module (DDCFM) is designed to facilitate effective interaction between phase and amplitude features, as well as between the frequency and spatial domains, enabling more comprehensive information aggregation. Extensive experiments on the IXI and fastMRI knee datasets demonstrate that PAS-Mamba consistently outperforms state-of-the-art methods under various acceleration factors and sampling patterns, validating the effectiveness and robustness of the proposed framework. 

Despite its effectiveness, the PAS-Mamba framework is limited to 2D MRI reconstruction without explicit modeling of inter-slice correlations. Extending PAS-Mamba to 3D reconstruction is a natural direction for future work and may further enhance reconstruction quality and clinical relevance.



\noindent\textbf{Declaration of competing interest}

The authors declare that they have no known competing financial interests or personal relationships that could have appeared to influence the work reported in this paper.

\noindent\textbf{Acknowledgements}

This work is supported by the National Natural Science Foundation of China (Nos.U22A2034, 62302530),  Major Program from Xiangjiang Laboratory (No.23XJ02005), Key Research and Development Programs of Department of Science and Technology of Hunan Province(No.2024JK2135), the Scientific Research Fund of Hunan Provincial Education Department (No.24A0018), Hunan Provincial Natural Science Foundation (No.2023JJ40769) and Central South University Research Programme of Advanced Interdisciplinary Studies(No.2023QYJC020).

\bibliography{mybibfile}

\end{document}